# Uncertainty Measurement of Deep Learning System based on the Convex Hull of Training Sets

Hyekyoung Hwang, Jitae Shin*, *Member, IEEE*

*Abstract*— Deep Learning (DL) has made remarkable achievements in computer vision and adopted in safety critical domains such as medical imaging or autonomous drive. Thus, it is necessary to understand the uncertainty of the model to effectively reduce accidents and losses due to misjudgment of the Deep Neural Networks (DNN). This can start by efficiently selecting data that could potentially malfunction to the model. Traditionally, data collection and labeling have been done manually, but recently test data selection methods have emerged that focus on capturing samples that are not relevant to what the model had been learned. They're selected based on the activation pattern of neurons in DNN, entropy minimization based on softmax output of the DL. However, these methods cannot quantitatively analyze the extent to which unseen samples are extrapolated from the training data. Therefore, we propose To-hull Uncertainty and Closure Ratio, which measures an uncertainty of trained model based on the convex hull of training data. It can observe the positional relation between the convex hull of the learned data and an unseen sample and infer how extrapolate the sample is from the convex hull. To evaluate the proposed method, we conduct empirical studies on popular datasets and DNN models, compared to state-of-the art test selection metrics. As a result of the experiment, the proposed To-hull Uncertainty is effective in finding samples with unusual patterns (e.g. adversarial attack) compared to the existing test selection metric.

## I. INTRODUCTION

Recently, Deep learning (DL) system has made remarkable progress in computer vision [1]–[3]. Since it learns based on huge amount of data with on their own and requires minimal human knowledge, the application of DL has been increased in many safety-critical scenarios, such as medical image analysis [4]–[6] and autonomous driving [7]–[9]. However, it showed several problems such as irrelevant misclassification with unfounded evidence when it applied to unlearned area [10]–[12]. This became one of the major factors of hindering its application in real-world. Thus, research purpose to figure out the decision base of DL [13], [14], calculate confidence [15], [16] or clarify an uncertainty [17], [18] of DL is one of the key contribution topics in these days.

It is obvious that test the trained model on every possible data is not available. In addition, to evaluate the trained model through prediction results on test data, we need a ground truth of the test set, which is mainly based on manual labeling. This is time consuming and often requires expertise to be precise. Therefore, it is very important to choose a test input worth labeling from a large, unlabeled input, with a limited labeling budget. Thus, research about developing a Test Priority Selection (TSP) [19] is getting a lot of attention since it purposes to find samples that are lack of relevance with training data, and it should reveal the uncertainty of the trained model upon unseen data. The adoption of this technology usually identifies the corner cases that reveal unlearned attributes. Therefore, it can provide informative candidate training samples upon the model.

However, developing a TSP metric for DL systems is challenging due to the complexity of the tasks and massive amount of data they learned. Recent studies [20], [22]–[29] have attempted to represent the novelty of the sample by making assumptions about what can represent relevance with the training data and then measuring it. Therefore, in order to understand test selection, it is important to understand how each study defined the 'lack of relevance' between the training data and the unseen sample.

First, there is Neuron Coverage (NC) [22]–[24], which introduced the traditional concept of code coverage [20] used in software testing to activation of DNN. Code coverage refers to the ratio of the code executed through testing among the source codes constituting the software. When it is applied to activation of DNN model are activated during the testing process and NC itself is regarded as the relevance with training set. This is recognized as the representative of test selection of deep learning system.

The next method deals a sample relevance with training set as the model's softmax output, assuming that it is a probability distribution formed by the learned decision boundary. This uses methods such as selecting a test sample that yields a model decision in the direction of decreasing entropy in the overall learning status [26] or selecting a sample whose softmax distribution is close to a uniform distribution [27]. In addition, by

This work was supported in part by a National Research Foundation of Korea (NRF) Grant funded by the Korean Government Ministry of Science and ICT (MSIT) under Grant NRF-2020R1F1A1065626, and in part by the MSIT under the Information Technology Research Center (ITRC) support program (IITP-2022-2018-0-01798) supervised by the Institute for Information \& Communications Technology Planning \& Evaluation (IITP).

The authors are with the Department of Electrical and Computer Engineering, Sungkyunkwan University, Korea.

Color versions of one or more of the figures in this article are available online at http://ieeexplore.ieee.org

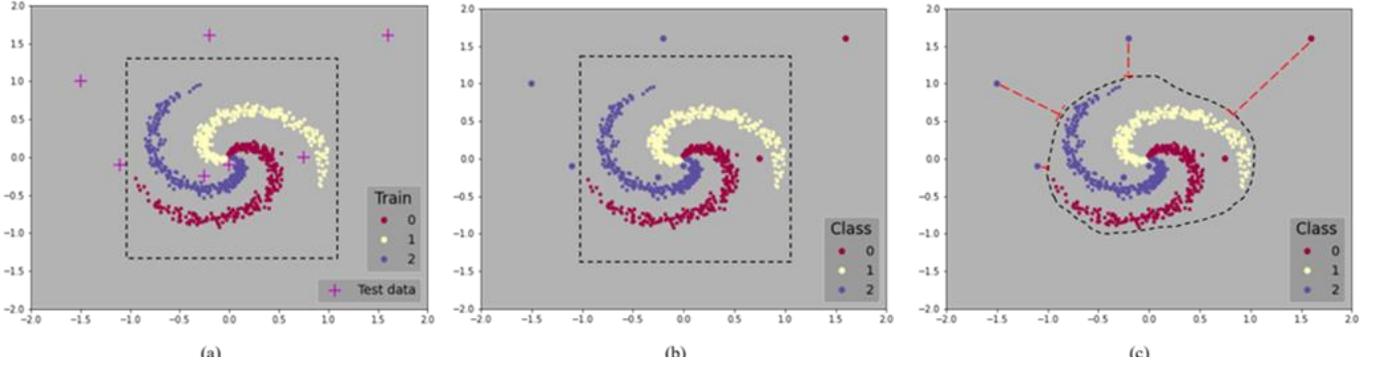

**Fig 1.** Example situation of test data positions with bounded training data. (a) shows training data with its range and test samples, (b) shows training data with its range and test samples prediction results, (c) shows training data with its convex hull and test samples with distance to the convex hull. Black dashed square implies the range of training set, black dashed curve is the convex hull of training set, Red dashed implies the distance the exterior and the convex hull.

adding dropout during testing, uncertainty is measured through the model's output variation (e.g., variance) for the same input [28], [29].

Also, there are studies that measure the relevance as how much a model 'surprise adequacy' for new inputs [25]. Here, surprise usually refers to the dissimilarity between the neuron activation value in the training data and the neuron activation value in the new input. It is similar to NC in that it uses the neuron activation value, but it is different in that added value of individual input can be evaluated.

However, why don't we consider the position of unseen data while selecting the test set? It goes without saying that you cannot learn on all data. This means that a range for the training data exists, so the test sample can or cannot exist within the range. Figure 1 provides an example of this situation. The dotted points in Figure 1(a) are training data and the magenta colored '+' marks are the test data. The training set is bounded by the black dashed square and test data exist inside and outside of the dashed square. In Figure 1(b), the magenta colored '+' marks in Figure 1(a) are now changed to its prediction result based on the trained classification model. At this time, samples lying outside of the dashed square are decided from extrapolation of the model decision boundary. In this case, are the uncertainties of the samples inside and outside of the training set range are the same? Or the uncertainty of a specific sample is independent of the degree of its extrapolation? Existing studies do not consider the geometrical relationship of test set and training data. Therefore, we propose to measure sample uncertainty based on the topological relationship between the test sample and the training set boundary.

Before figuring out the distance of sample extrapolation distance, it was necessary to consider how to measure the range of the training set. To solve the problem, we decided to introduce a convex hull to represent the range of training data, which is motivated by [34]. [34] examines the training set convex hull in pixel-level of classification task dataset from the algorithm of [35]. The dashed black diagram in Figure 1(c) shows the approximated convex hull of the training set. Therefore, the red dashed line represents the distance of each sample, which located outside of the convex hull, to the convex hull. According to Figure 1(c), the test sample in the upper right corner of the convex hull is the least relevant sample with training set. In addition, we also wanted to determine how much of the given test data is contained within the convex hull.

Therefore, this research estimates the convex hull of the training set according to the method of [35], and then proposes To-hull Uncertainty (TU), which quantifies an uncertainty according to the positional relationship of unseen samples with the convex hull. Our proposed TU is a metric for separate sample, and it helps to understand the intuitive interpretation of the sample location upon the convex hull. In addition, based on TU, we propose a Closure Ratio (CR) that can judge the status of the entire test set. This is an applicable indicator in the model selection.

The experimental results demonstrate that proposed TU shows best performance on detecting adversarial examples and guidance on adversarial attack and showed the possibility of using CR in model selection. Also, the mixture with the proposed TU and test selection metric with different viewpoints complemented the shortcomings from each viewpoint to show excellent performance. In summary, our contributions in this paper are:

- We propose a test selection metric with geometrical relation of unseen sample and training set convex hull, which is called To-hull Uncertainty (TU)
- Proposed TU shows compatible performance on distinguishing unusual samples (e.g., adversarial example)
- We propose a metric that shows the extrapolation status of test set upon training set, which is called Closure Ratio (CR), which can apply to model selection.
- The proposed metrics are task-agnostic, and in classification, it can be used together with other test selection metric.

The rest of this paper is organized as follows. Section II introduces the backgrounds of test selection, convex hull in deep learning as related work. Section III presents details of our two proposed metrics. Section IV presents experimental set ups to evaluate our proposed methods. Section V contains the analysis of experimental results, and we conclude the study in Section VI.

## II. BACKGROUNDS

The purpose of this paper is to identify and quantify how the degree of extrapolation of an unseen sample and calculate the uncertainty of the sample upon the model.

### A. Test Selection Metric of Deep Learning System

After DL has shown successful performance, testing neural networks according to conventional methods [20], [21] had emerged, which is called Neuron Coverage Criteria. DeepXplore [22] introduced the first white-box metric, Neuron Coverage (NC). It considers a neuron to be covered if the output of a neuron is greater than the parameter k, which is defined as the ratio of the number of covered neurons to the total number of neurons. They assumed that the higher the activation coverage, the more logic of more DNNs can be explored. Therefore, it was considered that the more difficult the test data was obtained as the NC was maximized, and it was shown that higher the NC value increased the L1 distance between inputs. A follow-up study, DeepTest [24], leveraged the same assumptions as [22] and it showed that different im- age transformations lead to different neuron coverage values to identify degenerate relationships that hold in specific contexts. DeepGauge [23] observed the NC in terms of the effectiveness of the test data and showed that the existing neuron coverage could not distinguish the clean test data from the adversarial attacked data. To solve this, they proposed several neuron coverage criteria with different granularity. For example, the Top-*k* NC calculated the coverage through *k* neurons that were most active in each layer and *k* Multi-section NC divided the interval of neuron activation values obtained in the train process into *k* equal sections and calculated the ratio of the covered sections to the total sections.

Later, [25] found that coverage criteria were impractical for selecting appropriate test data because they did not distinguish the added value of individual test inputs. They insisted that attributes suitable for test data selection should guide the selection of individual inputs, suggesting a surprise adequacy that measures how surprised the model is on new inputs. Here, surprise adequacy measures the dissimilarity of neuron activation values to new inputs compared to neuron activation values in training set. It was divided into two types according to how the dissimilarity of neuron activation was measured. The first one is distribution-based metric, called Likelihood Surprise Adequacy (LSA) and another one is distance-based metric, called Distance Surprise Adequacy (DSA).

Another test selection methods that differ from these are to use the softmax probability of the model. These are mainly work in classification task, assuming that the softmax output is the classification probability for each prediction produced by the learned decision boundary of the model. In [26], test data selection was performed using the softmax output value itself, and [27] selected samples that minimize cross entropy. Also, in [28], the Gini Index of each sample was calculated from the softmax output to measure the sample uncertainty. Furthermore, [29] pointed out the over-confident problem of the extrapolated sample in DL. To solve this problem, they applied Dropout [30] during the test to check the fluctuations of the decision boundary upon the sample so that the larger variation implies higher uncertainty of the sample upon the model.

### B. Test Selection Metric of Deep Learning System

In deep learning, the concept of convex hull has been mostly used to in neural network verification (NNV) [31] – [33]. NNV studied to verify the safety properties of models in a controlled environment by performing optimal convex abstraction on deep learning neurons. However, they have limitations in that they have to be tested in a strictly controlled environment, so it is difficult to apply to the real-world, and the computational cost is high.

However, new perspective of utilizing convex hull in deep learning have emerged. In [34], the algorithm to approximate convex hull in higher dimension [35] was used to investigate the convex hull of the most commonly used data in computer vision in pixels-level, and it showed that the test sample are mostly distributed outside the convex hull in a pixel-level. They also argue that extrapolation is necessary to improve the generalization performance of the model, and over-parametrization is necessary to achieve it. However, they neither investigate the status in feature-level nor quantified the over-parametrization degree. Inspired by this, we measure the degree of sample extrapolation on both feature and pixel level based on the convex hull and propose a metric that can represent the status of positional distribution of test set upon the convex hull.

## III. PROPOSED METRICS FOR UNCERTAINTY

In this section, we introduce details about our proposed To-hull Uncertainty, which utilize a distance between a sample and the convex hull of training set. Also, we introduce a data status inspection metric, Closure Ratio, that reveals the ratio of samples that are included to the convex hull among the test set.

### A. To-hull Uncertainty (TU)

Suppose that we have N samples of d-dimension training set $X \in R^{N \times d}$ and M samples with same dimension test set $X_T \in R^{M \times d}$. The i-th sample of $X_T$ is $X_{T_i}$. The convex hull of X is denoted as $C_X$. Then, $X_{T_i}$, which is a sample of $X_T$ or unseen data in general, have three possible positions to $C_X$: **Interior**, **Boundary** and **Exterior**. Thus, the closure sample indicates the sample lies in the interior or the boundary of $C_X$ and the exterior sample as placed in the exterior of $C_X$.

**Algorithm 1** Training set Convex Hull Point Set Approximation based on [35]
***
Initialize: Training set $X \in \mathbf{R}^{N \times d}$, $C_X = \emptyset$, $d = 0$, $i = 0$
1: Calculate $\epsilon > 0$
2: $U$ = Convex Hull of $X \leftarrow \emptyset$
3: $C_X \leftarrow x_0 \in X$
4: $X \leftarrow X - \{x_0\}$
5: WHILE ($i \leq N$):
   $i \leftarrow i + 1$
   $x_i = \text{argmax}_X Distance(X, C_X)$
   {Choose $x_i \in X$ which are farthest from $C_X$}
   $d \leftarrow Distance(t_i, C_X)$
   IF ($d \leq \epsilon$):
      return $C_X$
   ELSE:
      $C_X \leftarrow C_X \cup t_i$
      $X \leftarrow X - \{t_i\}$
Output: $C_X$
***

We utilize the algorithm from [35] that approximates the convex hull in high dimension by sparse point set generation. The proof and details of the algorithm can be found in [35]. Since they proposed a general algorithm, we apply the algorithm to the deep learning circumstances.

The algorithm for point set approximation of training set convex hull is shown in Algorithm 1. The algorithm was proposed to approximate a point set that conducts convex hull in higher dimension. Therefore, we would like to show a two-dimensional example to help understand the result of the algorithm. Figure 2 shows the example of estimated point set constituting the convex hull of the toy data used in Figure 1. For the simplicity, the training data are shown as gray dots without class distinction, and the black '×' mark represent the approximated point set that consists of $C_X$. The left side of Figure 2 represents the set of approximated points set of $C_X$ with it's blue-dashed line margin boundary and the right side shows how we calculate the sample distance to $C_X$. Here, dashed line is for $D(X_T, P_{X_T})$. Note that this algorithm is a point set approximation, not finding the vertices of an exact convex hull.

Since conditions such as dimension of data and number of samples are different for each dataset, it is not suitable to use a fixed value as the margin. Therefore, we set adaptive $\epsilon$ for each data rather than using a fixed value. At this time, to create a convex hull that contains the properties of the training set, the margin should be calculated only with the attributes of the training set, and the value should be less affected by outliers. Thus, the average of self-excluded minimum pairwise distance is set as $\epsilon$ shown in (1)

$$\epsilon = \frac{1}{N} \cdot \Sigma_{i=1}^{N} \min_{\forall x_j \neq x_i} \sqrt{(x_i - x_j)^2}, \quad x_j, x_i \in X. \quad (1)$$

To figure out whether a specific sample is included in the $C_X$. or not is determined through the $\epsilon$. Remind the termination condition of the $C_X$ approximation algorithm, the element of X that is the furthest from $C_X$ also has a distance to $C_X$ less than or equal to $\epsilon$. This means that all training data have a distance to $C_X$ less than or equal to $\epsilon$. Therefore, if the distance to $C_X$ of a sample $X_{T_i}$ is less than $\epsilon$, the sample is a closure sample and vice versa.

To calculate the distance between the $X_{T_i}$ and $C_X$, notated as $D(X_{T_i}, P_{T_i})$, $X_{T_i}$ is first projected to $C_X$ and this projection point is called $P_{T_i}$. Then, the Euclidean distance between the $P_{T_i}$ and $X_{T_i}$ represents $D(X_{T_i}, C_X)$, which is shown below

$$D(X_{T_i}, C_X) = \sqrt{\Sigma_{j=1}^{d}(X_{T_i} - P_{T_i})^2}. \quad (2)$$

With $D(X_{T_i}, C_X)$ and the margin of $C_X$, which is $\epsilon$, the proposed To-hull Uncertainty (TU) is computed as (3)

$$To - hull\ Uncertainty = \frac{D(X_{T_i}, C_X)}{\epsilon}. \quad (3)$$

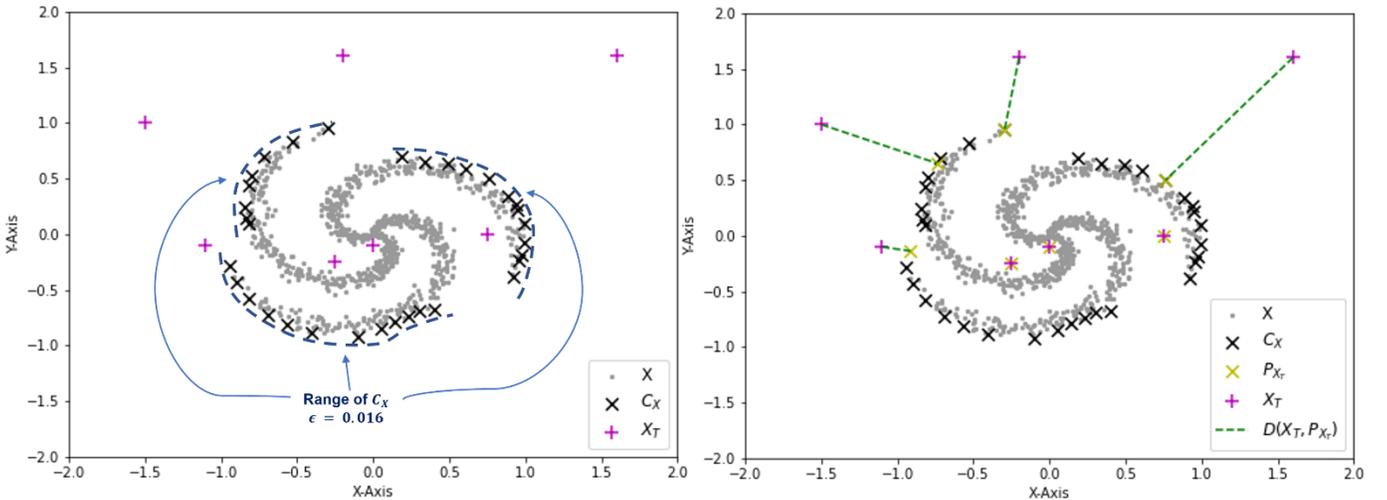

yellow '×' mark represent the projection point $P_{X_T}$ and green-

**Fig 2.** Example of estimated point set based on [35], constituting the convex hull of the toy data in Figure 1. **Left:** the approximated point set of training data convex hull with it's margin, $\epsilon$. **Right:** $P_{X_T}$ on the approximated point set convex hull and $D(X_T, P_{X_T})$.

| Dataset | | Description | Model | # Params (MB) | # Layers | # Original Test Images | # Adversarial Test Images for each attack |
|---|---|---|---|---|---|---|---|
| MNIST | | Handwritten Digits 0~9 classification | SmallCNN | 4.58 | 4 | 10000 | 10000 |
| CIFAR-10 | | Images with 10 class classification | ResNet18, | 42.63 | 19 | 10000 | 10000 |
| | | | ResNet50 | 89.72 | 51 | | |
| CIFAR-100 | | Images with 100 class classification | ResNet18, | 42.8 | 19 | 10000 | - |
| | | | ResNet50 | 90.43 | 51 | | |
| MedMNISTv2 | Pneumonia | Chest X-ray with binary (0/1) classification | - | - | - | 624 | - |
| | Retina | Fundus Images for 5-level ordinal regression | - | - | - | 400 | - |
| | Dermatology | Dermato-scope Images for 7 class classification | - | - | - | 2005 | - |

TABLE I
DATASET AND MODEL DESCRIPTIONS FOR OUR EXPERIMENT DESIGN. WE APPLY MEDMNISTV2 TO PIXEL-LEVEL EXPERIMENT ONLY.

TABLE II
DETAILS OF SMALL CNN FOR MNIST

| Layer | Size of Kernel | Input Size | Output Size |
|---|---|---|---|
| Conv2D | 3x3 | (1, 28,28) | (32, 26, 26) |
| ReLU | - | (32, 26, 26) | (32, 26, 26) |
| Conv2D | 3x3 | (32, 26, 26) | (64, 24, 24) |
| ReLU | - | (64, 24, 24) | (64, 24, 24) |
| MaxPool | 2x2 | (64, 24, 24) | (64, 12, 12) |
| Flatten | - | (64, 12, 12) | 9216 |
| Linear | - | 9216 | 128 |
| ReLU | - | 128 | 128 |
| Linear | - | 128 | 10 |

If $X_{T_i}$ is a closure sample of $C_X$, then $0 \leq TU(X_{T_i}) \leq 1$ and if it is an exterior sample, then $1 < TU(X_{T_i})$. Also, as the sample is further away from $C_X$, $TU(X_{T_i})$ increases. For example, if $TU(X_{T_a}) = 2$, $TU(X_{T_b}) = 4$, then the $X_{T_a}$ lies in $2 \cdot \epsilon$ times farther to $C_X$ and $X_{T_b}$ lies in $4 \cdot \epsilon$ times farther. Thus, we can say that $X_{T_b}$ is farther from $C_X$ than $X_{T_a}$.

*B. Closure Ratio (CR)*

Based on the TU that provides the positional status of single data, an inspection of the entire dataset status is possible. Thus, we propose the Closure Ratio (CR), which implies the ratio of closure samples in $X_T$ compared to the convex hull $C_X$, shown in (4)

$$Closure\ Ratio = \frac{|M_{closure}|}{|M|}, \qquad M_{closure} \subseteq M \quad (4)$$

where $|M|$ and $|M_{closure}|$ is the cardinality of $M$ and the set of closure sample in $M$. Thus, the proposed CR is the ratio of samples with $0 \leq TU(X_{T_i}) \leq 1$ among the test set. If the closure ratio is 0.7, then 70% of $X_T$ are closure samples of $C_X$. Therefore, the larger the closure ratio, the more test samples are included in the training set convex hull.

IV. EXPERIMENT DESIGN

In this section, we describe the experimental setup, such as the dataset and model we used, and the hyper-parameters used to train the model, and approaches for adversarial image generation. Also, we introduce the latest test selection metric to compare with the proposed method and write three research questions and how the experiment was designed to confirm it. To conduct the experiments, we implement our approach as well as other methods upon PyTorch 1.12. All experiments were performed on a Ubuntu 20.04.5 LTS server with NVIDIA RTX 2080Ti GPU.

*A. Datasets and Models*

As shown in Table I, for evaluation, we select four popular publicly available datasets, i.e., MNIST [36], CIFAR-10 and CIFAR-100 [37], MedMNISTv2 [38]. The MNIST dataset is for handwritten digits recognition, containing 70,000 input data in total, of which 60,000 are training data and 10,000 are test data. The CIFAR-10 and CIFAR-100 dataset consists of 60,000 32x32 color images in 10 and 100 classes, with 6,000 and 600 images per class respectively. It consists of 50,000 train images and 10,000 test images. MedMNISTv2 is a MNIST-like dataset collection of standardized biomedical images, including 12 datasets for 2D and 6 datasets for 3D. Thus, all dataset has 28 x 28 resolution medical images. We select pneumonia, retina, dermatology dataset among 2D. Pneumonia is binary classification for gray-scale chest X-ray, which consists of 4,708 / 524 / 624 images for train, validation, test set. Retina is color image with ordinal regression, 1,080 / 120 / 400 for train, validation, test. Dermatology is color images for multi-class classification, which contains 7,007 / 1,003 / 2,005 for train, validation, test respectively.

The size of utilized CNN models ranges from tens to thousands of neurons, exhibiting the diversity of the models to some degree. For MNIST, we trained a small CNN with two convolution layers followed by max-pool, two fully connected layers, which is described in Table II, called SmallCNN with 99.18% accuracy. For CIFAR-10, ResNet18 [39] with accuracy 94.79% and ResNet50 [39] with 95.23% as fixed model, and for CIFAR-100, ResNet18 with accuracy 75.81% and ResNet50 with accuracy 78.15% are used. The final layer output of each model, which is an input of softmax layer is used to calculate DSA [25] and TU. For MedMNISTv2, we apply our proposed method in pixel level to demonstrate flexible applicability of our metric.

*B. Adversarial Test Input Generation*

In case of CIFAR-100, the experiment for prioritizing original tests is available since the accuracy of trained models are in the middle of 70s. However, other dataset, such as MNIST or CIFAR-10, we have to conduct an adversarial example prioritizing experiment because of the high performance of fixed models. We use four most common methods to generate adversarial tests, including Projected Gradient Descent (PGD) [40], Fast Gradient Sign Method (FSGM) [41], Carlini & Wagner (C&W) [42], Basic Iterative Method (BIM) [43]. These techniques generate tests through different minor perturbations on a given test input. The final column of Table I shows the number of adversarial tests generated by each adversarial attack methods.

*C. Comparison Approaches*

We did not compare Neuron Coverage series, which cannot be differentiated for individual added values. Instead, we compare our proposed TU with DSA [25], which calculates the dissimilarity of neuron activation values based on Euclidean distance, and DeepGini [28], which is the latest test selection technique based on the softmax value.

*D. Research Questions*

TU is designed for quantifying the extrapolation distance of the test samples from the training set. This is a new approach based on a geometric perspective, unlike the traditional methods that are focus on malfunctions of the model. Therefore, we have composed three research questions (RQ) to verify the proposed metric.

**RQ1. Sample Novelty:** Is proposed TU can captures the relative difference of an input of DL system?

We provide answers to RQ1 in two ways. Firstly, we visualize the test data that has high TU and low TU values and compare whether the higher TU valued sample shows unusual pattern.

Secondly, we evaluate whether it is possible to capture the adversarial sample based on proposed metric. As [25] done, we generate 10,000 adversarial images for each adversarial attack methods based on original test set of CIFAR-10 and MNIST. Then, calculate the test selection metric of total 20,000 images. Finally, train a logistic classifier based on the test selection metrics of randomly selected 1,000 original and 1,000 adversarial inputs and evaluate the trained classifier with the test selection metrics of 18,000 images. If the values of metrics correctly capture the variation of DL system's behavior, the trained logistic classifiers should detect adversarial examples successfully. We report accuracy for evaluation since it utilizes all values in confusion matrix of classification.

**RQ2. Correlation:** Is TU correlated to existing test selection metrics in DL?

Since our proposed metric has different perspective on lack of relevance than others, we want to examine the correlation of our proposed metric with others.

To provide answers of RQ2, we made two types of correlation coefficient. To figure out the linearity between the TU and existing methods [25, 28], we calculate the Pearson correlation coefficient [44] between them. Also, we Point-biserial correlation coefficient [45] between TU and classification result (correct=1, wrong=0) to measure the correlation between TU and correctness of the unseen sample.

**RQ3. Guidance:** Can TU guide retraining of DL systems to improve their accuracy against test set or adversarial examples?

To evaluate the guidance of TU, we made two types of experiments. The accuracy of the CIFAR-100 on the test set is usually in the middle of 70%. So, after retraining the model including up to n% of the test set, we check the model accuracy for the remaining (100-n)% ($n \leq 10$).

However, in the case of CIFAR-10 or MNIST, since the DL accuracy for the test set is more than 95%, it is difficult to confirm the guidance for the original test set. Therefore, we confirm the retrain guidance for the test set to which the aforementioned four adversarial attacks are applied. This is the model accuracy for the remaining (100-n)% adversarial attacked test set after additional 50 epochs training by creating a new training set including n% ($n \leq 10$) adversarial attacked test set in the original training set for the baseline model.

V. PERFORMANCE ANALYSIS

In this section, we present the results of each research question and then analyze whether our geometrical perspective of sample uncertainty estimation is available to capture the samples with unusual pattern and contains consistent result with existing methods.

*A. Status Analysis based on TU and CR*

In this subsection, we provide how to analyze the status of data and model in DL based on our proposed metrics. Table III shows the summary of our proposed metric on the dataset and models we used. Each row represents a proposed metric. In the case of TU, it represents the average value for the exterior sample to check how far the exterior samples are from the training data according to the state of each data. Each column represents data to which the suggested metric is applied. And 'Pixel' column indicates the result of applying the proposed metric in the pixel space of the corresponding dataset, and the column indicated by the name of model such as 'ResNet18' represents the result of applying the proposed metric in the trained model's feature space of the dataset. The information of the model used, and the accuracy of the trained model are described in detail in Section IV.

TABLE III
SUMMARY OF TU AND CR ON THE DATASET & MODELS. BOLD FOR COMPARISON OF PIXEL AND FEATURE-LEVEL, UNDERLINED FOR THE LEAST UNCERTAINTY CONDITION BASED ON TU AND CR

| Dataset | | | CIFAR-10 | | | CIFAR-100 | | | Derma | Retina | Pneumonia |
|---|---|---|---|---|---|---|---|---|---|---|---|
| | Pixel | SmallCNN | Pixel | ResNet18 | ResNet50 | Pixel | ResNet18 | ResNet50 | Pixel | Pixel | Pixel |
| **CR** | 0.272 | **<u>0.937</u>** | 0.048 | **0.911** | **<u>0.936</u>** | 0.46 | **<u>0.564</u>** | **0.47** | 0.51 | 0.4 | 0.26 |
| **TU(exterior)** | 1.288 | **<u>1.22</u>** | 1.89 | **1.43** | **<u>1.32</u>** | 1.31 | **<u>1.25</u>** | **1.27** | 1.24 | 1.36 | 1.42 |

According to Table III, the feature level shows higher CR and lower TU than the pixel level, as expressed in bold in each data. In the case of MNIST, the CR is 0.048 at the pixel level, but all have values greater than 0.9 in the feature space. In the case of TU, the pixel level was 1.89, but at the feature level, it was reduced to 1.43 and 1.32. This means that the inclusion of unseen samples for the training set convex hull is increased by mapping the high-dimensional input to the low-dimensional through DL. In addition, the average distance for the training set convex hull of the exterior sample also decreased from 1.89 times to 1.43 times and 1.32 times depending on the model, respectively. These analyzes can also be applied to model selection and are underlined in the table. In the case of CIFAR10, ResNet18 shows higher CR and lower TU than ResNet50, so it can be supported that selecting ResNet50 is a better decision in the evaluation based on sample inclusion relationship. In CIFAR-100, ResNet18 outperforms ResNet50 in this viewpoint.

*B. RQ1: Sample Novelty*

In this section, we provide an experiment result for RQ1 and its analysis to check whether the proposed TU can find an unusual type of samples compared to the training data.

*1) Sample Visualization:* Figure 3 is a visualization of randomly selected samples within the TU range according to the TU value of each dataset at the pixel level. We divide the range of visualization in three ranges. The left side visualized test samples are the closure to $C_X$ so that the value of TU is greater than or equal to 0 and less than or equal to 1. The figures placed in the middle and on the right are exterior test samples for each data $C_X$. The samples located in the middle are the exterior samples with TU values greater than 1 and less than or equal to 2. The right-side shows samples with TU values greater than 2, which are the farthest samples among the exterior samples. According to the Figure 3, among the exterior samples, the patterns of the images are very different from that of the closure as the TU increases. In the case of medical images, pictures with problems such as FLARE phenomenon (Fundus) or error of angle of view (Pneumonia) are also observed. In the case of handwriting, noticeably anomaly written letters are shown.

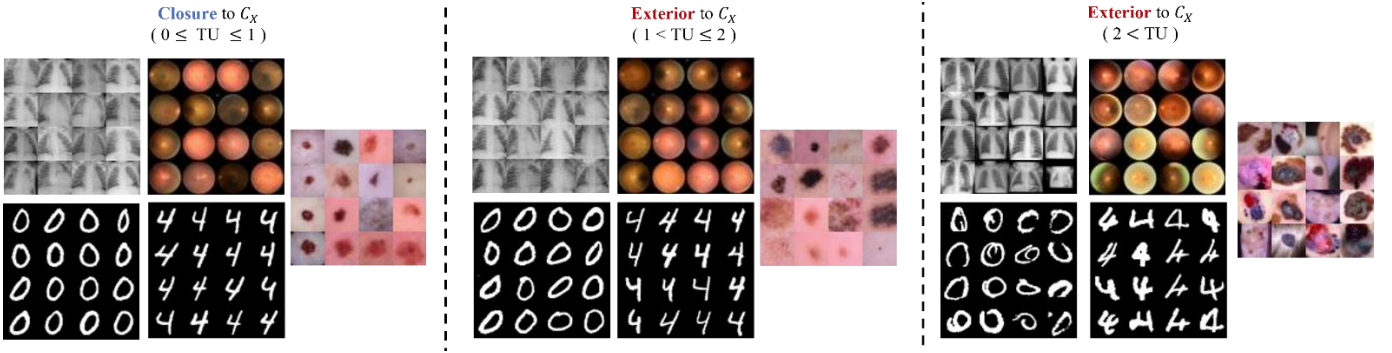

**Fig 3.** Visualization of samples in MedMNISTv2 and MNIST based on the **Pixel-level TU**. The left-side shows the closure test samples, $0 \leq TU \leq 1$. The figures placed in the middle and on the right are exterior samples for each $C_X$. The middle ones show the exterior samples with TU values greater than 1 and less than or equal to 2. The right-side shows samples with TU values greater than 2, which are the farthest samples among the exterior samples.

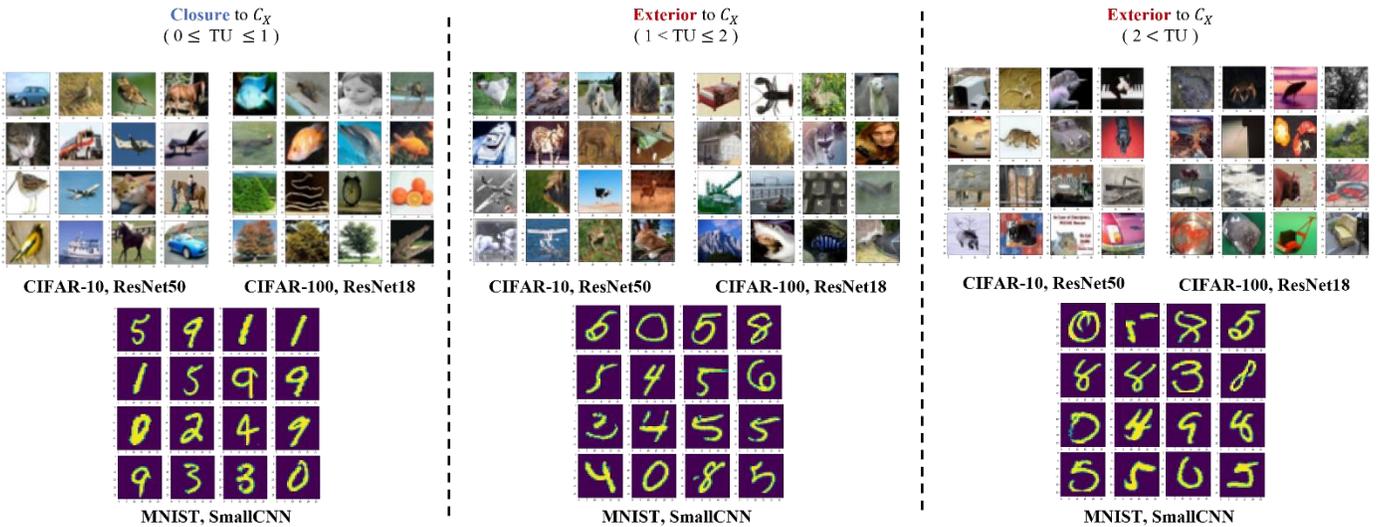

**Fig 4.** Visualization of samples in CIFAR-10, CIFAR-100, based on the **Feature-level TU**. The left-side shows the closure test samples, $0 \leq TU \leq 1$. The figures placed in the middle and on the right are exterior samples for each $C_X$. The middle ones show the exterior samples with TU values greater than 1 and less than or equal to 2. The right-side shows samples with TU values greater than 2, which are the farthest samples among the exterior samples.

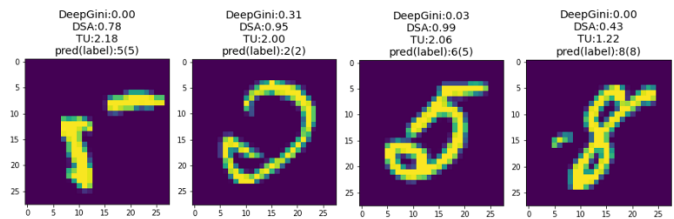

**Fig 5.** Samples with high TU at the feature level with other test selection metrics [25], [28]

Figure 4 is a visualization of randomly selected samples according to the TU value of each dataset at the feature level. As in Figure 3, it is divided into three ranges. The left side visualized test samples are the closure to $C_X$ so that the value of TU is greater than or equal to 0 and less than or equal to 1. The figures placed in the middle and on the right are exterior test samples for each data $C_X$. The middle picture shows the exterior samples with TU values greater than 1 and less than or equal to 2. The right-side shows samples with TU values greater than 2, which are the farthest samples among the exterior samples. The name of each dataset and the model utilized is written below each image. Similar to the pixel level result, if the TU value is increased at the feature level, we can observe the distorted or disconnected handwriting in MNIST, the partial objects, the images with text in background that are not related to objects, or the image with awkward background in CIFAR dataset.

Figure 5 shows samples with high TU at the feature level with other test selection metrics. According to the figure, only TU gives high uncertainty about samples that show abnormal shapes while others doesn't. In case of DeepGini [28], which calculates the sample uncertainty based on softmax output, it cannot give a low uncertainty if the model makes predictions with high probability no matter it is correct or not. In the case of DSA, it shows higher uncertainty than DeepGini, but it is lower than the proposed metric, and it cannot be interpreted as intuitively as the suggested TU.

*2) Adversarial Image Classification:* Adversarial attack confuses the model by perturbing the input image. Therefore, distinguishing them from the model's point of view plays an important role in preserving performance. Therefore, to check whether the proposed metric can discriminate the sample novelty caused by adversarial attack, we created a logistic classifier that distinguishes the two with the test selection metric of a clean sample and an adversarial image. A higher performance of the trained classifier indicates that it can distinguish sharply between the clean and adversarial images of each image.

TABLE IV
PERFORMANCE OF LOGISTIC CLASSIFIER TRAINED ON **TSP METRIC** OF 1,000 CLEAN AND 1,000 ADVERSARIAL SAMPLE, EVALUATED ON THE METRIC OF REST 9,000 CLEAN, 9,000 ADVERSARIAL IMAGES, RESPECTIVELY.

| **Average Accuracy (%)** | **TU(F)** | DSA | DeepGini |
|---|---|---|---|
| **MNIST** | **88.18** | 77.42 | 75.30 |
| **CIFAR-10** | **84.17** | 72.14 | 72.08 |

TABLE V
PERFORMANCE OF LOGISTIC CLASSIFIER TRAINED ON **COMBINED TSP METRIC (TSP METRIC · [28])** OF 1,000 CLEAN AND 1,000 ADVERSARIAL SAMPLE, EVALUATED ON THE METRIC OF REST 9,000 CLEAN, 9,000 ADVERSARIAL IMAGES, RESPECTIVELY.

| **Average Accuracy (%)** | **TU(F)** | DSA | DeepGini |
|---|---|---|---|
| **MNIST** | **96.05** | 90.27 | 80.55 |
| **CIFAR-10** | **95.82** | 82.22 | 75.03 |

We represent the performance of the learned logistic classifier on Table IV. Each column represents a sample novelty measure metric that utilizes to train the logistic classifier. TU at the feature level is written as TU(F), and TU at the pixel level is written as TU(P). The results show that the proposed metric at the feature level has at most 9% higher accuracy to discriminate adversarial samples compared to the conventional metric in both data. This indicates that using TU is more efficient than other methods when checking for contamination of data.

In addition, since the proposed metric has different viewpoint of uncertainty from others, TU can be used together with the metric proposed in other studies. For example, an application on adversarial samples classification with combination of TU and DeepGini is shown in Table V. Even though the performance of logistic classifier trained with $DSA \cdot DeepGini$ shows gain compared to the classifier trained only with $DSA$, the classifier trained with $TU \cdot DeepGini$ shows the best performance. This is because DeepGini compensates for the closure sample in the proposed metric but ambiguous in the decision boundary of the model, and the TU compensates the exterior sample, but the model is over-confident.

*C. RQ2: Correlation*

In this section, we provide an experiment result for RQ 2 and its analysis to check whether the proposed TU has correlation with existing test selection method and decision correctness of model.

TABLE VI
POINT-BISERIAL CORRELATION COEFFICIENT OF TEST SELECTION METRICS AND MODEL PREDICTION CORRECTNESS

TABLE VII
POINT-BISERIAL CORRELATION COEFFICIENT OF TEST SELECTION METRICS AND MODEL PREDICTION RESULT

TABLE VIII
RETRAIN GUIDANCE OF TSP METRICS ON CIFAR-10 ADVERSARIAL IMAGE

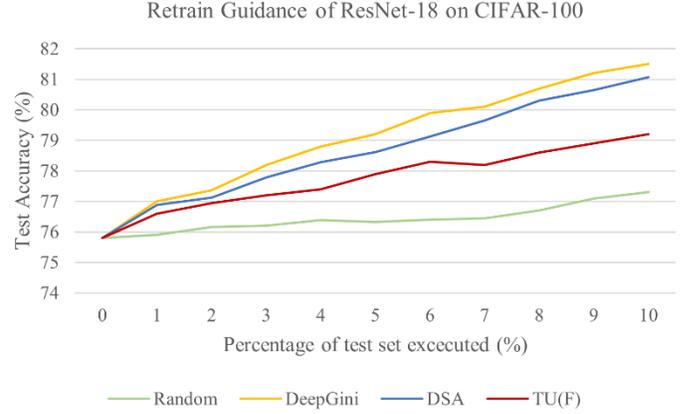

**Fig 6.** Retrain Guidance of each TSP metric on CIFAR-100 original test set.

The proposed TU only considers the uncertainty of unseen samples for the training data and the distance from the training set convex hull. Therefore, we measured the Point-biserial correlation coefficient [45] with the success of prediction in the CIFAR-10 and CIFAR-100 experimental environments to check whether the prediction success of the model is related to TU. In Table VI, the proposed TU shows low correlation compared to the existing metric, but still has weak correlation between model prediction correctness. This is because the existing methods include information related to whether the model prediction is suited or not, such as softmax output or calculate the distance of training sample that has same classification with test sample, while TU simply measures the distance. This lower correlation can be analyzed the natural phenomenon that if the sample is far from the trained pattern that model has been learned, then the model could be wrong to the sample.

Additionally, we want to check whether our perspective has a similar correlation with existing studies. Therefore, we measured the Pearson correlation coefficient [44] with compared metrics in the CIFAR-10 and CIFAR-100 experimental environments to check whether our perspective is relevant with others. The correlation between TU and existing TSP metric is shown in Table VII. The proposed TU in feature level has a correlation with DSA, which is measured based on the Euclidean distance between unseen data and train samples in feature space, rather than DeepGini that using softmax probability. Therefore, in the area where DSA can be utilized, feature level TU can also be applied.

### D. RQ3: Retrain Guidance

In this section, we evaluate the retrain guidance of proposed metric in two ways: Retrain guidance on the original test set and adversarial image. As mentioned in Section IV, to demonstrate the efficiency of data utilization, we only applied less than 10% additional data. Also, to minimize the effect on the random state, we experimented by changing the random seed 5 times and report the average of them.

*1) Retrain Guidance on Original Test Set:* The evaluation of retrain guidance efficiency on original test set is made on CIFAR-100 since the model's accuracy on the data is below 80%. The result is shown in Figure 6. The x-axis shows the percentage of the added test set for retraining, and the y-axis shows the retrain accuracy at that time. The point where the x-axis is 0 is the baseline, and the accuracy of the ResNet18 model in baseline is 75.81%.

According to Figure 6, the accuracy for the original test set gradually increases as percentage of additional test set increases even in random selection. The result is consistent result with point-biserial correlation coefficient in RQ2. When retraining the model by gradually adding 1%, ... , 10% of the test set to the existing training set, sample selection using TU at the feature level shows an efficient increase in accuracy compared to random selection but lower than existing metrics. This is because existing methods are focused to sample the mis-classified images from the baseline for retraining, while ours concentrated on how the sample is extrapolated from the convex hull of training set.

*2) Retrain Guidance on Adversarial Test Set:* To observe the retrain guidance on adversarial image, we retrain the model with adding n% ($n \leq 10$) of the adversarial attacked test set to the existing clean training set and calculate the classification accuracy for adversarial attacked images of (100-n)%. To avoid the effect of random state, we experimented by changing the random seed 5 times and report the average of them. Higher retrain guidance indicates better defense efficiency against adversarial samples.

TABLE IX
RETRAIN GUIDANCE OF TSP METRICS ON MNIST ADVERSARIAL IMAGE

| Attack Type | Metrics | 1% | 2% | 3% | 4% | 5% | 6% | 7% | 8% | 9% | 10% |
|---|---|---|---|---|---|---|---|---|---|---|---|
| BIM | DeepGini | 66.49 | 63.77 | 71.26 | 73.92 | 66.73 | 66.93 | 70.65 | 70.95 | 67.67 | 71.81 |
|  | DSA | 68.7 | 63.42 | 70.44 | 71.21 | 75.44 | 71.9 | 72.59 | 72 | 73.88 | 77.76 |
|  | **TU(F)** | **72.03** | **73.81** | **76.38** | **79.19** | **80.16** | **79.04** | **82.66** | **83.32** | **83.76** | **85.39** |
| PGD | DeepGini | 64.03 | 66.63 | 66 | 72.42 | 72.34 | 72.13 | 70.65 | 71.17 | 73.08 | 70.84 |
|  | DSA | 66.56 | 68.84 | 70.82 | 70.74 | 72.99 | 73.12 | 74.69 | 74.89 | 73.38 | 74.54 |
|  | **TU(F)** | **70.02** | **76.17** | **77.09** | **79.24** | **81.29** | **81.24** | **83.23** | **84.55** | **83.89** | **85.88** |
| C&W | **DeepGini** | **93.25** | **96.28** | 95.93 | **97.14** | 96.32 | **96.94** | 96.63 | 97.01 | **97.02** | **97.39** |
|  | DSA | 94.6 | 94.28 | 95.95 | 95.79 | 96.24 | 96.62 | 96.45 | 97.02 | 96.93 | 97.06 |
|  | TU(F) | **94.97** | 95.45 | **96.28** | 96.16 | **96.79** | 96.49 | **96.85** | **97.13** | **97.29** | **97.56** |
| FGSM | DeepGini | 74.03 | 78.31 | **82.33** | 83.38 | 84.52 | 85.26 | **87.73** | 86.2 | 90.64 | 90.33 |
|  | DSA | **75.36** | **79.61** | 80.24 | 81.52 | 82.67 | 84.81 | 86.66 | 88.58 | 89.25 | 89.49 |
|  | **TU(F)** | 74.74 | 78.87 | 80.34 | **83.91** | **85.99** | **86.76** | 87.16 | **89.55** | **91.65** | **90.84** |

The result of evaluation on retraining guidance for adversarial attack in CIFAR-10 for ResNet18 and ResNet50 is shown in Table VIII. For simplicity, average the value of retraining guidance accuracy for 1$\sim$10% in here. According to the former description, the average accuracy is 72.8% when repeating 1% to 10% retrain guidance for C&W adversarial attack with feature level TU with 5 different random seeds. Thus, each cell in Table VIII is the average value of 50 experiments. From the result, the proposed TU shows the best performance, means that retraining the ResNet18 with additional 10% adversarial attacked CIFAR-10 images that has highest feature-level TU with original test set, the expected accuracy on the corrupted images is 68.687%, 67.019%, 68.343%, 73.937% for each PGD, FGSM, BIM, C\&W attack, respectively.

The quantitative result of the retraining guidance for adversarial attack in MNIST is shown in Table IX. In this case, we report all accuracy for adversarial attacked image of (100-n)%, ($n \leq 10$). It means that if the SmallCNN is retrained by adding data whose TU value is the top 1% among BIM attacked samples to the original test set, it shows an average classification accuracy of 72.03% for the remaining 99% of BIM attacked data. The best performance in each n% is shown in bold, and the metric that performed the best performance the most out of 50 experiments is written in bold. As a result, adversarial sample selection using TU at the feature level showed an efficient increase in accuracy compared to other metrics.

VI. CONCLUSION

In this paper, we propose a new perspective on relevance of training set and unseen sample that considers the positional relation of the training set convex hull and the sample. Based on it, we proposed a sample uncertainty quantification metric, called To-hull Uncertainty, which measured from the distance to the convex hull, by approximating the training set convex hull. Also, we propose the Closure Ratio for model selection and data status analysis. To evaluate our metric, we set up three research questions and conducted verification on them. As a result, the proposed metric showed the best performance in discriminating adversarial images and retrain guidance on adversarial images while it shows a weak correlation with the correctness of the model. For further work, we intend to verify the test selection metric for both epistemic uncertainty, which is model uncertainty, and aleatoric uncertainty, which is data uncertainty. Also, we'll update the convex hull approximation with a better algorithm such as [46].